\documentclass{article}
\usepackage{spconf,amsmath,graphicx}
\usepackage{comment}
\usepackage{verbatim}
\usepackage{amsmath}
\usepackage{amssymb}
\usepackage{amsmath}
\usepackage{amssymb}
\usepackage{booktabs}
\usepackage{multirow}
\usepackage{bbding}
\usepackage{textcomp}
\usepackage{tablefootnote}
\usepackage[accsupp]{axessibility}
\usepackage{hyperref}
\usepackage{graphicx}
\usepackage{subcaption}
\usepackage{enumitem}
\hypersetup{pagebackref,breaklinks,colorlinks}

\title{CHUG: Crowdsourced User-Generated HDR Video Quality Dataset}
\twoauthors
  {Shreshth Saini, Alan C. Bovik}
	{The University of Texas at Austin \\
	Austin, TX, USA}
  {Neil Birkbeck, Yilin Wang, Balu Adsumilli}
	{YouTube, Google Inc.\\
	Mountain View, CA, USA}

\begin{document}
%\ninept
%

\maketitle

\begin{abstract}  
High Dynamic Range (HDR) videos enhance visual experiences with superior brightness, contrast, and color depth. The surge of User-Generated Content (UGC) on platforms like YouTube and TikTok introduces unique challenges for HDR video quality assessment (VQA) due to diverse capture conditions, editing artifacts, and compression distortions. Existing HDR-VQA datasets primarily focus on professionally generated content (PGC), leaving a gap in understanding real-world UGC-HDR degradations. To address this, we introduce \textbf{CHUG}: Crowdsourced User-Generated HDR Video Quality Dataset, the first large-scale subjective study on UGC-HDR quality. CHUG comprises 856 UGC-HDR source videos, transcoded across multiple resolutions and bitrates to simulate real-world scenarios, totaling 5,992 videos. A large-scale study via Amazon Mechanical Turk collected 211,848 perceptual ratings. CHUG provides a benchmark for analyzing UGC-specific distortions in HDR videos. We anticipate CHUG will advance No-Reference (NR) HDR-VQA research by offering a large-scale, diverse, and real-world UGC dataset. The dataset is publicly available at:~\url{https://shreshthsaini.github.io/CHUG/}.  
\end{abstract}

\begin{keywords}
Crowdsourced, High Dynamic Range (HDR), Video Quality Assessment, HDR VQA Dataset, User-Generated Content (UGC)
\end{keywords}

\begin{figure*}[htbp]
    \centering
    \includegraphics[width=\textwidth]{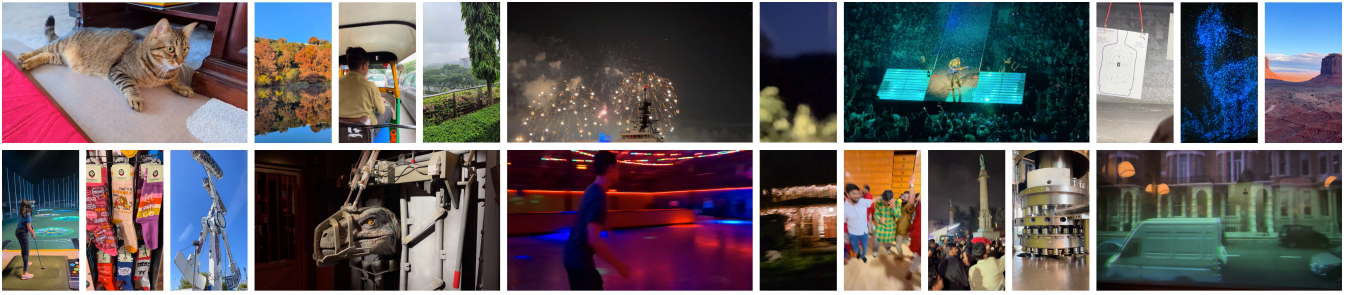}
    \caption{Sample frames from the CHUG dataset, showcasing diverse real-world UGC-HDR content with variations in lighting, motion, orientation, and distortions. Best viewed when zoomed in.}
    \label{fig:dataset-sample}
\end{figure*}

\section{Introduction}
\label{sec:intro}
The rapid growth of User-Generated Content (UGC) on platforms such as YouTube, Instagram, and TikTok has transformed digital media consumption~\cite{omnicore_tiktok, 99firms_facebook, mohsin2020youtube}. 
\begin{table}[htbp]
\small
\centering
\caption{Overview of the CHUG Dataset. The dataset comprises diverse UGC-HDR videos across multiple resolutions and bitrates, with extensive subjective quality annotations.}
\label{tab:dataset_overview}
\begin{tabular}{p{3.2cm} p{4.2cm}}
\hline
\textbf{Attribute} & \textbf{Details} \\
\hline
\multicolumn{2}{c}{\textbf{Video Specifications}} \\
\hline
Format & Rec. 2020, 10-bit, PQ \\
Resolutions & 1920$\times$1080, 1080$\times$1920, 1280$\times$720, 720$\times$1280, 640$\times$360, 360$\times$640\\
Duration & 4 - 10 sec. \\
\hline
\multicolumn{2}{c}{\textbf{Dataset Statistics}} \\
\hline
Reference Videos & 856 (428 Portrait, 428 Landscape) \\
Total Videos & 5,992 \\
Total Scores & 211,848 \\
Avg. Scores per Video & 35 \\
\hline
\end{tabular}
\end{table}
UGC is characterized by diverse capture conditions, user-editing effects, and platform-specific compression, which complicates Video Quality Assessment (VQA)~\cite{patchvq, konvid}. Simultaneously, High Dynamic Range (HDR) is gaining adoption due to its wider color gamut, higher bit depth, and enhanced luminance, enabling improved perceptual quality~\cite{cta2024,ITU_BT2100_2018}. However, UGC-HDR VQA remains an open problem due to HDR-specific distortions such as banding, tone-mapping artifacts, and exposure non-uniformity~\cite{hdrchipqa,hidrovqa}. Existing HDR datasets such as LIVE-HDR~\cite{livehdr} and SFV+HDR~\cite{sfvhdr} focus on professionally generated content (PGC) and lack scale and diversity to represent real-world UGC scenarios. LIVE-HDR~\cite{livehdr} contains only 31 reference HDR videos with professionally captured content. The ITM-HDR-VQA dataset~\cite{itm-hdr-vqa} offers 200 inverse tone-mapped HDR videos, while the KVQ dataset~\cite{kvq} and SFV+HDR dataset~\cite{sfvhdr} focus on portrait short-form content. To the best of our knowledge, there exists no large-scale, publicly available UGC-HDR dataset that is representative of real-world HDR distortions. CHUG addresses these gaps by providing the largest UGC-HDR dataset with real-world distortions, authentic content diversity, and high-quality subjective scores, serving as a benchmark for advancing NR-VQA models. The key characteristics of this dataset are summarized in Table~\ref{tab:dataset_overview}. The key contributions are:
\begin{itemize}[noitemsep,leftmargin=*]
    \item CHUG includes 856 UGC-HDR source videos, transcoded across resolutions and bitrates, creating 5,992 videos—the largest HDR and UGC-HDR dataset to date.
    \item CHUG applies a bitrate ladder encoding strategy, replicating social media compression on UGC-HDR videos.
    \item The first large-scale HDR study on Amazon Mechanical Turk (AMT), collecting 211,848 ratings, with each video rated by 35 subjects on average.
    \item CHUG serves as a benchmark for UGC-HDR distortions, aiding NR-VQA model development for real-world HDR streaming.
\end{itemize}

%CHUG provides a diverse, realistic UGC benchmark for developing and evaluating HDR-VQA models for social media streaming platforms.

\section{Crowdsourcing UGC-HDR Videos}
\label{sec:video_collection}
In this section, we discuss the construction of a large-scale UGC-HDR video quality dataset by collecting real-world HDR videos from users. To ensure diverse content, orientations, and distortions, we curated the dataset and applied bitrate ladder encoding to simulate real-world streaming conditions.

\subsection{Source Video Collection and UGC Diversity}
\label{sec:source_collection}

The dataset was built through an open call for UGC-HDR video submissions, where contributors uploaded HDR videos from personal devices such as iPhones, Samsung Galaxy, Google Pixel, and OnePlus smartphones. All submissions included appropriate consent and rights transfer agreements. We applied strict filtering criteria to all submitted videos to ensure dataset integrity by removing duplicates, objectionable content, and static or minimally dynamic videos. Each video was trimmed to a maximum 10 seconds using \texttt{ffmpeg}~\cite{ffmpeg}, with no resolution upscaling or downscaling beyond 1080p to preserve original quality.

%\textbf{DOUBTFULL/REMOVE}\textit{To ensure CHUG aligns with real-world UGC distributions, we extracted spatial and temporal features following \cite{patchvqa}, including absolute luminance, colorfulness, RMS contrast, and face detection statistics. The dataset’s feature distribution was validated against YouTube-UGC, KoNViD, and LSVQ ~\cite{youtubeUGC,KoNViD,LSVQ}, confirming its statistical alignment with large-scale UGC datasets.}

\begin{table}[htbp]
\small
\centering
\caption{Bitrate ladder used for dataset creation. Each video was encoded at multiple bitrates to simulate real-world streaming conditions.}
\label{tab:bitladder}
\begin{tabular}{p{3.5cm} p{3.5cm}}
\hline
\textbf{Resolution} & \textbf{Bitrates (Mbps)} \\
\hline
360p  & 0.2 \\
720p  & 0.5, 2.0 \\
1080p & 0.5, 1.0, 3.0 \\
\hline
\textbf{1080p} & Reference \\
\hline
\end{tabular}
\end{table}
% --- Footnote appears here outside the float ---
%\footnotetext{Based on YouTube's streaming guidelines \cite{youtube_bitrate} and Apple's HLS authoring specifications \cite{apple_hls}.}

\begin{figure}[htbp]
    \centering
    \includegraphics[width=0.25\textwidth]{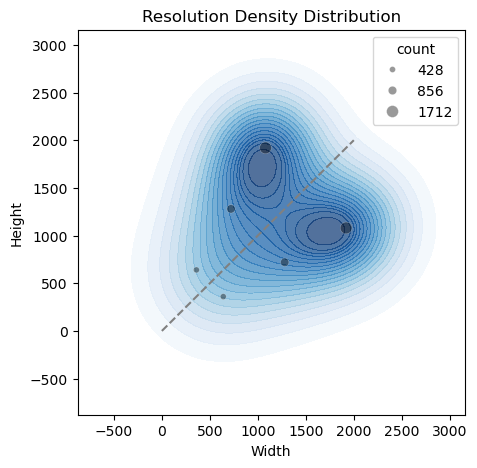}
    \caption{Resolution distribution of CHUG dataset, maintaining a balanced mix of landscape and portrait videos to study orientation-based perceptual differences.}
    \label{fig:resolution}
\end{figure}

An equal mix of landscape and portrait videos was maintained to analyze orientation-based quality perception. Fig.~\ref{fig:resolution} presents the resolution distribution of CHUG, although different resolution has different number of videos but for single resolution there is an equal split of landscape and portrait videos. The dataset spans urban environments, natural landscapes, indoor vlogs, and sports recordings, covering a range of lighting conditions, including daylight, nighttime, and extreme brightness scenarios, see Fig.~\ref{fig:dataset-sample}.

\begin{table*}[htbp]
\centering
\small
\caption{Comparison of CHUG with existing HDR VQA datasets.}
\label{tab:dataset_comparison}
\begin{tabular}{lccccc}
\hline
\textbf{Dataset} & \textbf{Format} & \textbf{Total Videos(Ref.)} & \textbf{Source} & \textbf{Total opinions} & \textbf{Orientation} \\
\hline
LIVE-HDR & Rec. 2020, HDR10, PGC & 310 (31) & Internet Archive & 20,400 & Landscape \\
SFV+HDR(\textit{only HDR}) & Rec. 2020, HDR10, UGC & 300 & YouTube & N/A & Portrait \\
CHUG (Ours) & Rec. 2020, HDR10, UGC & 5,992 (856) & Crowdsourced & 211,848 & Portrait/Landscape \\
\hline
\end{tabular}
\end{table*}

\subsection{Bitrate Ladder for Realistic Streaming Simulation}
\label{sec:bitladder}
UGC videos uploaded to social media platforms undergo platform-specific compression and transcoding, introducing bitrate-dependent distortions. To replicate these effects, we applied a bitladder encoding strategy~\cite{livehdr}, following YouTube’s streaming guidelines~\cite{youtube_bitrate} and Apple’s HLS authoring specifications~\cite{apple_hls}. Table~\ref{tab:bitladder} details the bitladder used in CHUG, ensuring that each video was encoded at multiple bitrates to simulate real-world streaming conditions.

\begin{figure}[htbp]
    \centering
    \includegraphics[width=0.8\linewidth]{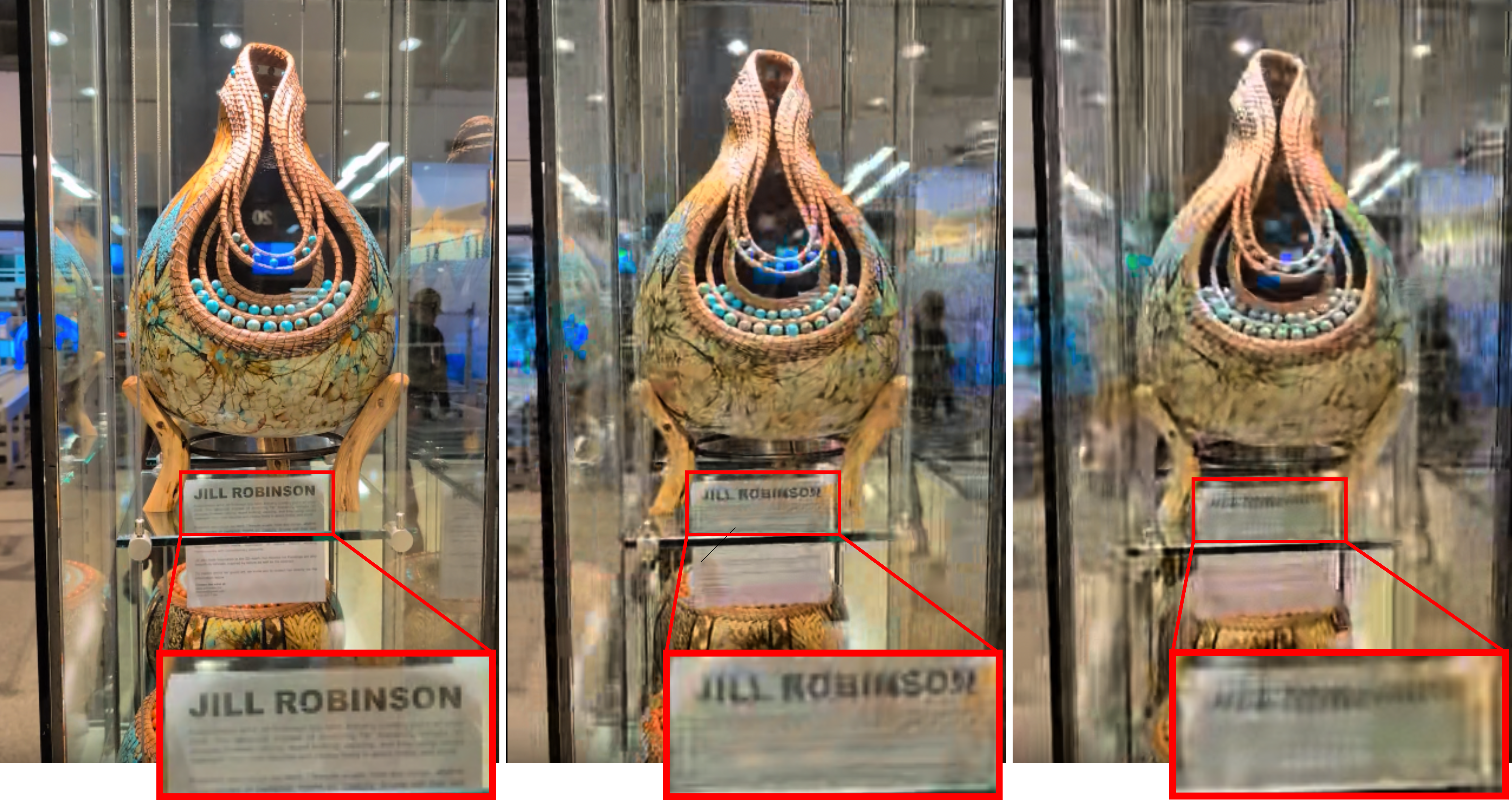}
    \caption{Compression artifacts introduced via bitladder. Left: 1080p reference, Middle: 720p at 2 Mbps, Right: 360p at 0.2 Mbps.}
    \label{fig:compression_artifacts}
\end{figure}

This encoding process introduced controlled compression artifacts, allowing us to analyze how bitrate variation impacts subjective HDR video quality. Fig.~\ref{fig:compression_artifacts} illustrates visible compression artifacts at different resolutions. The leftmost frame represents the original 1080p reference video, the middle frame is 720p at 2 Mbps, and the rightmost frame is 360p at 0.2 Mbps, showing noticeable degradation due to aggressive compression. The combination of authentic UGC-HDR content and real-world streaming simulation makes CHUG a challenging benchmark dataset.

%----------------------------------------------------

\section{Details of Subjective Study}
\label{sec:study}
In this section, we describe the subjective study conducted using Amazon Mechanical Turk (AMT)~\cite{mturk} to gather perceptual opinion scores for UGC-HDR videos. This is the first large-scale AMT-based UGC-HDR study, overcoming remote HDR evaluation challenges such as device compatibility, display limitations, and network constraints. To ensure robust data collection, strict filtering criteria were implemented, resulting in 211,848 ratings from 700+ subjects, with each video receiving an average of 35 ratings. 

\begin{figure}[htbp]
    \centering
    % Subfigure 1
    \begin{subfigure}[b]{0.5\textwidth}
        \centering
        \includegraphics[width=0.6\textwidth]{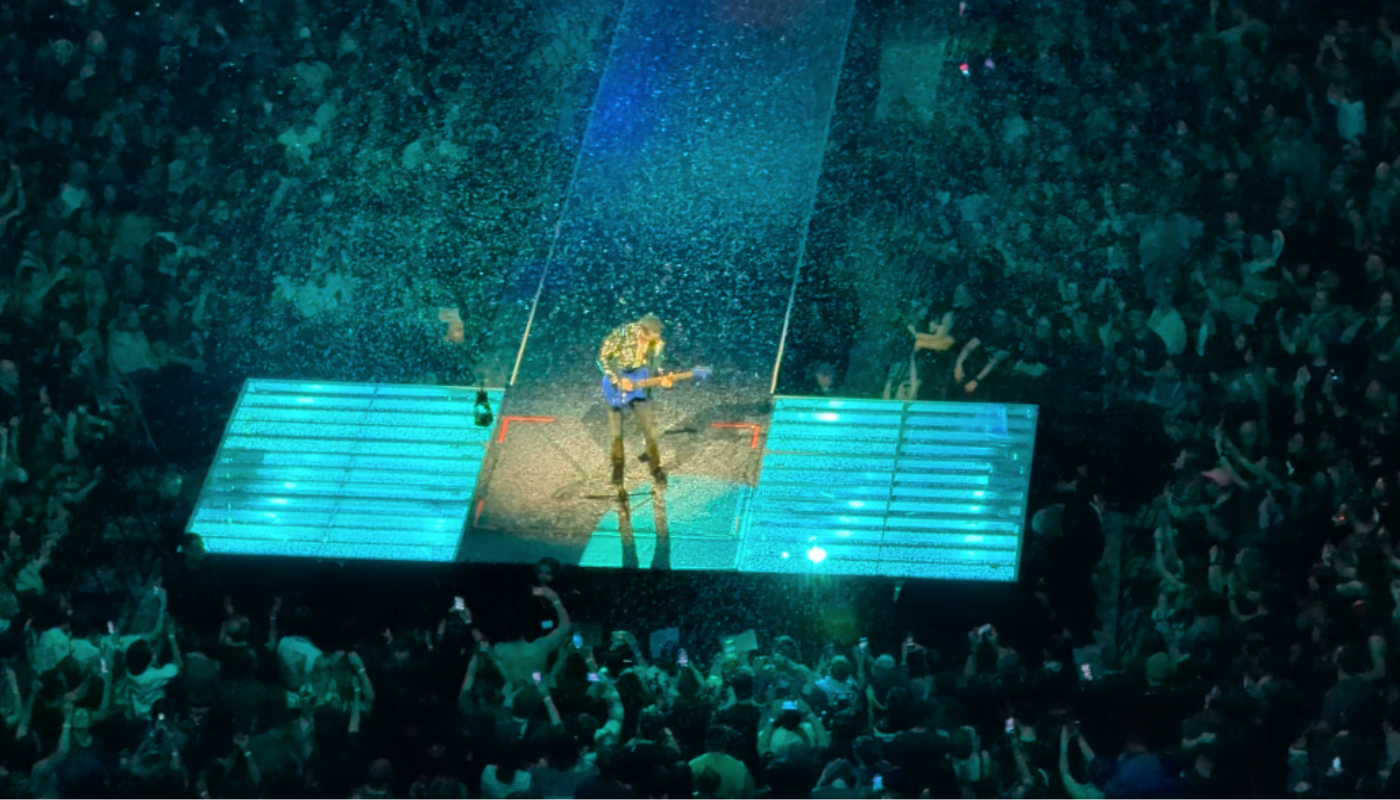}
    \end{subfigure}
    \hspace{0.5em} % Small horizontal space between figures
    % Subfigure 2
    \begin{subfigure}[b]{0.5\textwidth}
        \centering
        \includegraphics[width=0.9\textwidth]{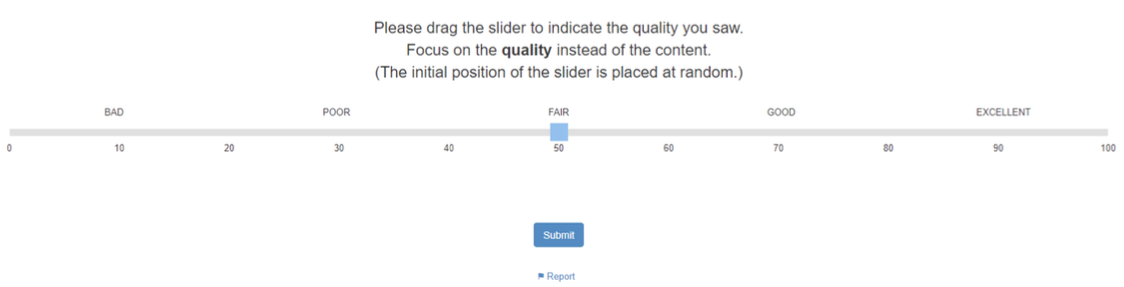}
    \end{subfigure}
    \caption{Rating interface used for the AMT study (Best viewed zoomed in).}
    \label{fig:amt_study}
\end{figure}

\subsection{Study Design and Rating Procedure}
Each Human Intelligence Task (HIT) pipeline began with instructions and a brief quiz to confirm participant understanding. Subjects then completed a training phase by rating six HDR videos, familiarizing themselves with the rating interface~, see Fig.~\ref{fig:amt_study}. The testing phase involved rating 94 videos, each presented only once, on a 0-100 Likert scale. Videos were pre-loaded to prevent buffering, and participants could flag videos for technical or content issues. At the end of each session, subjects provided device specifications, demographics, and viewing conditions. A pilot study was conducted with experienced AMT users to establish a golden set of 1428 videos, later used for participant reliability filtering.

\subsection{Participant Screening and Quality Control}
To ensure data reliability, multiple subject and session rejection criteria were applied~\cite{patchvq,paqtopiq}, categorized as:\\
\textbf{Pre-screening:} Participants with incompatible devices (e.g., non-HDR displays) were blocked from proceeding. We dynamically checked bit depth, HEVC codec support, display resolution, and network speed.\\
\textbf{Training Phase:} Continuous HDR validation was performed to detect any device setting changes mid-task. Playback completion tracking was enforced to prevent participants from skipping through videos.\\
\textbf{Testing Phase:} Progressive quality checks were applied at 25\%, 50\%, and 75\% task completion. Subjects were removed if $>$50\% of their ratings were flagged for playback issues or if they exhibited inconsistent rating behavior.\\
\textbf{Post-study validation:} Of the 94 test videos, 10 were control videos (5 duplicates+5 golden set). Subjects with deviations exceeding 20-25\% on repeated/golden videos were excluded.

After subject rejection, a rigorous data cleaning process was implemented to ensure the reliability of collected ratings. First, all ratings from disqualified subjects or those experiencing over 50\% playback issues were removed. Second, responses from participants who reported not wearing necessary vision correction were excluded. Finally, we applied ITU-R BT.500-14~\cite{ITU_BT500-14} filtering criteria, leading to the removal of 60 additional subjects. After these refinements, the cleaned dataset contained 211,848 high-quality ratings, ensuring robust subjective evaluation.

\section{Data Analysis}
\label{sec:data-analysis}
In this section, we analyze the collected subjective scores to assess the quality distribution, inter-subject agreement, and variations in MOS based on video properties such as resolution, bitrate, orientation, and content complexity. We further compare CHUG against LIVE-HDR and SFV+HDR, highlighting its diversity and representation of real-world UGC-HDR quality.

\begin{figure}[htbp]
    \centering
    % Subfigure 1
    \begin{subfigure}[b]{0.2\textwidth}
        \centering
        \includegraphics[width=\textwidth]{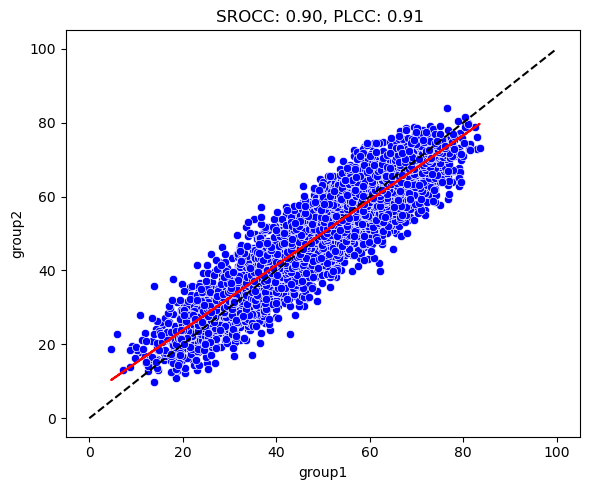}
        \caption{Inter-subject correlation}
        \label{fig:subfig1}
    \end{subfigure}
    \hspace{0.5em} % Small horizontal space between figures
    % Subfigure 2
    \begin{subfigure}[b]{0.2\textwidth}
        \centering
        \includegraphics[width=1.1\textwidth]{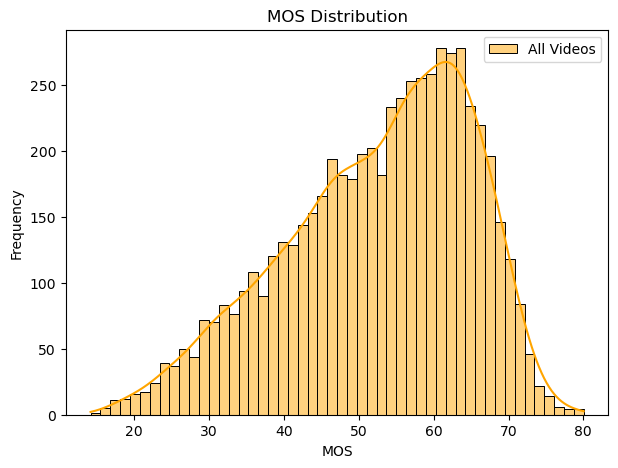}
        \caption{MOS distribution}
        \label{fig:subfig2}
    \end{subfigure}
    \caption{(a) MOS distribution of all videos; (b) Inter-subject correlation}
    \label{fig:mos_distribution}
\end{figure}

\subsection{Processing of Subjective Scores}
To obtain reliable Mean Opinion Scores (MOS), we employed the SUREAL (Subjective Reliability) method~\cite{sureal}, a robust statistical approach that accounts for subject biases and inconsistencies. Unlike traditional MOS calculations, which simply average subjective ratings~\cite{ITU_BT500-11}, SUREAL computes a Maximum Likelihood Estimate (MLE) of the true video quality, making it robust to outliers and unreliable subjects. The opinion scores $S_{i_j}$ given by subject $i$ for video $j$ were modeled as:

\begin{equation}
S_{ij} = \psi_{j} + \Delta_{i} + \nu_{i} X, \quad X \sim \mathcal{N}(0,1)
\end{equation}

Here, $\psi_{j}$ represents the true quality of video $j$. $\Delta_{i}$ accounts for the rating bias of subject $i$. $\nu_{i}$ models the inconsistency of subject $i$, ensuring that subjects with erratic rating behavior have less influence. The parameters $\psi_{j}$, $\Delta_{i}$, and $\nu_{i}$ were estimated using the Newton-Raphson optimization method to maximize the log-likelihood of observed scores. %This approach significantly improves the reliability of MOS, as it effectively reduces the impact of inconsistent raters and rating noise. %By applying SUREAL, we ensured that the final MOS values accurately reflected perceptual quality, making our dataset highly robust for benchmarking and video quality assessment.

\begin{figure}[htbp]
    \centering
    \includegraphics[width=0.6\linewidth]{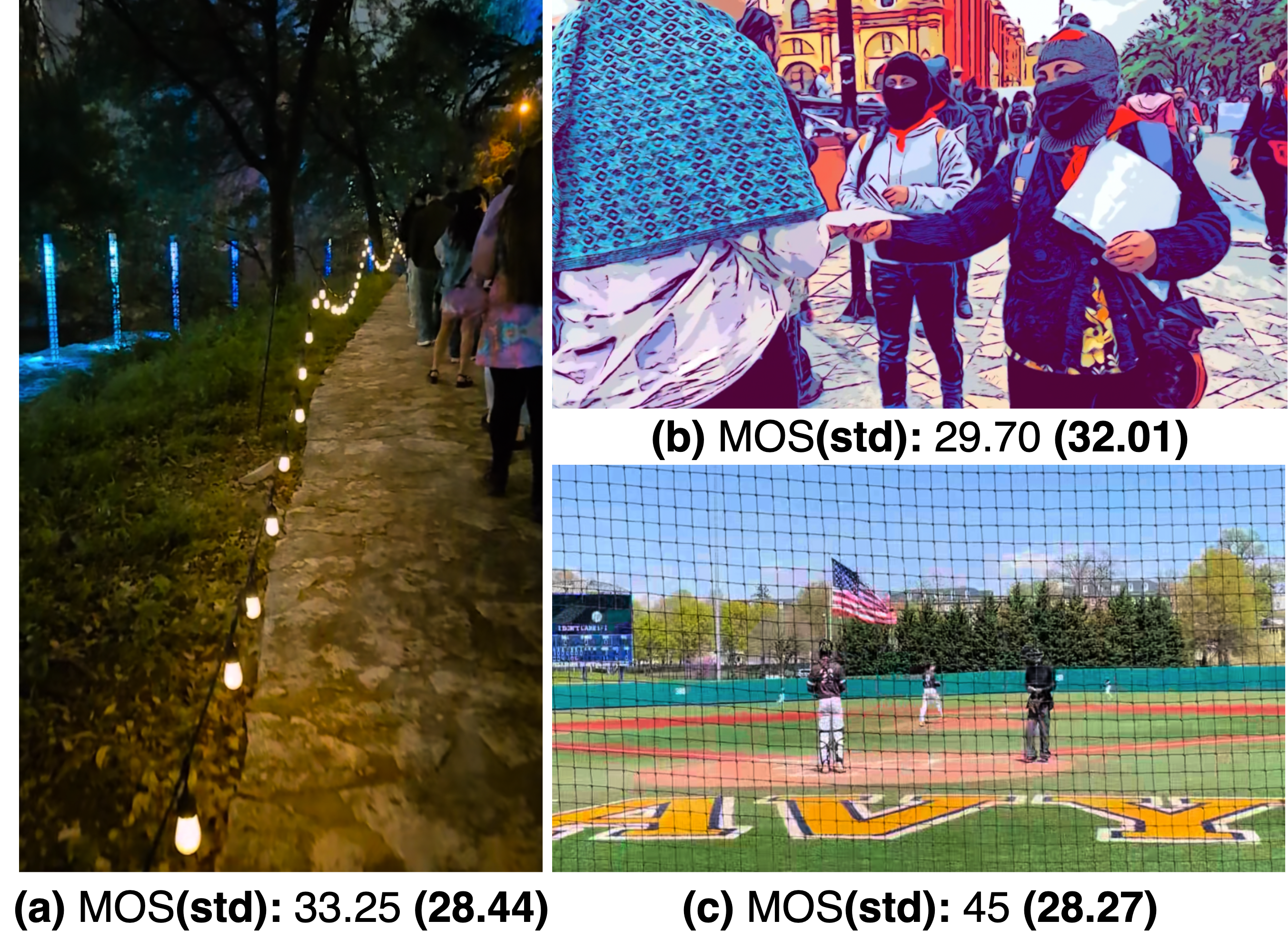}
       \caption{Examples of challenging videos with high MOS standard deviation. Night-time scenes, extreme filters, and occlusions caused rating inconsistencies.}
    \label{fig:challenging_videos}
\end{figure}

\subsection{Challenging Content and High Variance MOS}
Certain videos in CHUG exhibited high MOS variance, reflecting significant subjective disagreement among raters. Fig.~\ref{fig:challenging_videos} highlights examples of such challenging content, where night scenes, extreme filters, obstructions, and complex lighting conditions led to inconsistent ratings. These variations often result in a higher standard deviation in MOS, making them difficult for both human evaluators and objective quality assessment models. A major strength of CHUG is its inclusion of such real-world challenging scenarios, which are often missing from existing HDR datasets like LIVE-HDR~\cite{livehdr}, where professionally captured videos tend to have uniform lighting, controlled exposure, and minimal distortions. In contrast, CHUG incorporates UGC-HDR content from diverse real-world settings, making it a more versatile and realistic benchmark for NR-VQA models.

\begin{figure}[htb]
    \centering
    % Subfigure 1
    \begin{subfigure}[b]{0.2\textwidth}
        \centering
        \includegraphics[width=0.85\textwidth]{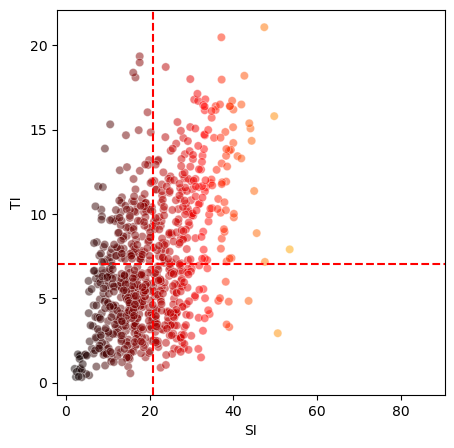}
        \caption{SITI}
        \label{fig:siti}
    \end{subfigure}
    \hspace{0.01em} % Small horizontal space between figures
    % Subfigure 2
    \begin{subfigure}[b]{0.2\textwidth}
        \centering
        \includegraphics[width=0.85\textwidth]{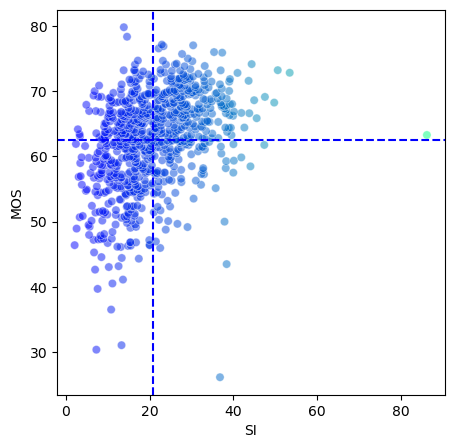}
        \caption{MOS vs. SI}
        \label{fig:mos_si}
    \end{subfigure}
    \hspace{0.01em} % Small horizontal space between figures
    \begin{subfigure}[b]{0.2\textwidth}
        \centering
        \includegraphics[width=0.85\textwidth]{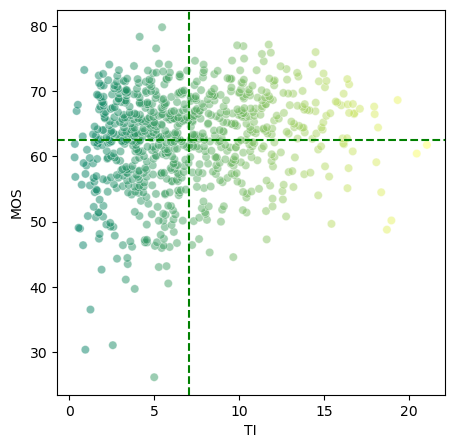}
        \caption{MOS vs. TI}
        \label{fig:mos_ti}
    \end{subfigure}
    \caption{(a) Spatial-Temporal Complexity, (b) MOS vs. Spatial Information (SI), and (c) MOS vs. Temporal Information (TI).}
    \label{fig:siti_mos}
\end{figure}

\subsection{Spatial-Temporal Features vs. MOS}
We analyzed the relationship between spatial (SI) and temporal (TI) complexity and perceived video quality. Fig.~\ref{fig:siti} visualizes the joint distribution of SI and TI across the dataset. Most videos cluster in the mid-to-high SI range with low-to-moderate TI, aligning with common UGC-HDR content characteristics. The spread of SI and TI values highlights CHUG’s diversity in motion and texture complexity. Fig.~\ref{fig:mos_si} indicates a positive correlation between SI and MOS, suggesting that videos with higher spatial details, such as textures and sharp edges, tend to receive higher quality ratings. This trend aligns with perceptual expectations.% as sharper videos generally appear more visually appealing. 
However, we also observe that extremely high SI values do not necessarily lead to the highest MOS, possibly due to compression artifacts becoming more noticeable in highly detailed regions. Fig.~\ref{fig:mos_ti} shows a non-monotonic trend between TI and MOS. While moderate motion complexity is associated with higher MOS, excessive motion (high TI) often results in lower ratings. This degradation is primarily due to motion compression artifacts, where aggressive encoding leads to blurring, ghosting, or unnatural frame interpolation effects. %Conversely, very low TI values (static scenes) also do not always yield the highest MOS, as such content may appear less engaging or lifelike. %This analysis underscores the interplay between spatial-temporal features and perceptual quality, emphasizing the challenges of modeling UGC-HDR content with varying levels of motion and detail.

\begin{figure}[htb]
    \centering
    % Subfigure 1
    \begin{subfigure}[b]{0.2\textwidth}
        \includegraphics[width=0.93\textwidth]{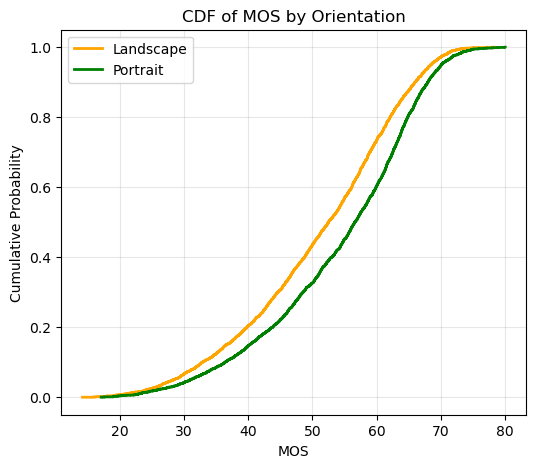}
        \caption{MOS CDF for orientations}
        \label{fig:orientation_cdf}
    \end{subfigure}
    \hspace{0.01em} % Small horizontal space between figures
    % Subfigure 2
    \begin{subfigure}[b]{0.2\textwidth}
        
        \includegraphics[width=1.2\textwidth]{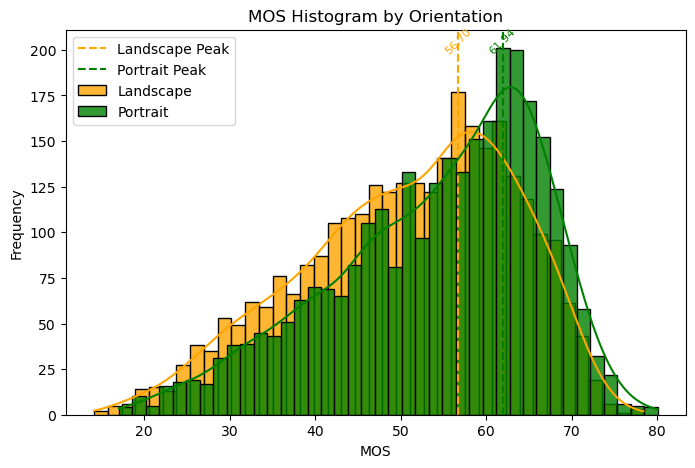}
        \caption{MOS distribution by orientation}
        \label{fig:orientation_distribution}
    \end{subfigure}
    \caption{MOS analysis for landscape vs. portrait videos.}
    \label{fig:orientation_mos}
\end{figure}

\subsection{Effect of Orientation on MOS}
Figure~\ref{fig:orientation_mos} analyzes the impact of video orientation (landscape vs. portrait) on perceived quality. Fig.~\ref{fig:orientation_distribution} shows a high degree of overlap between landscape and portrait MOS distributions, indicating that orientation alone does not significantly influence perceptual quality. Both orientations exhibit a similar MOS range and distribution shape, suggesting that factors such as content type, motion, and compression artifacts play a more dominant role in quality perception than orientation. Fig.~\ref{fig:orientation_cdf} provides a finer comparison using CDF curves. The curves indicate that portrait videos tend to receive slightly higher MOS scores, particularly in the mid-to-high quality range. %This may be attributed to portrait videos being more commonly associated with social media platforms where users engage with high-quality smartphone-recorded content optimized for HDR playback. %While the overall MOS distributions are similar, the subtle MOS advantage for portrait videos suggests that UGC-HDR content designed for mobile-first consumption may be better optimized for perceptual quality. However, the relatively small gap emphasizes that both orientations are equally important for HDR-VQA model development.

\begin{figure}[htb]
    \centering
    % Subfigure 1
    \begin{subfigure}[b]{0.2\textwidth}
        \includegraphics[width=1.1\textwidth]{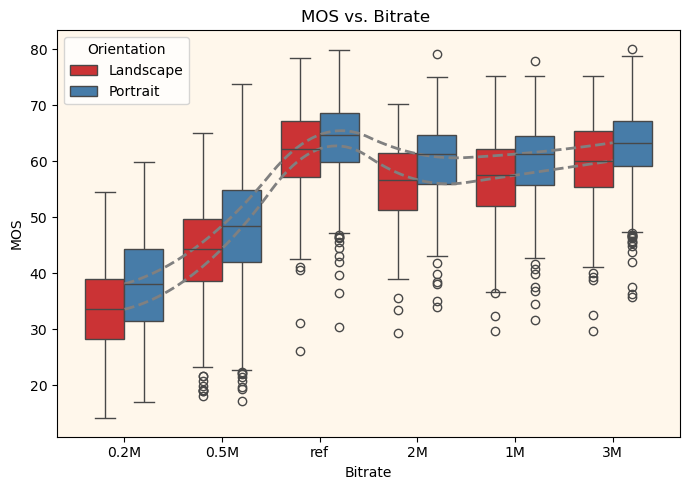}
        \caption{MOS vs. Bitrate}
        \label{fig:mos_bitrate}
    \end{subfigure}
    \hspace{0.5em} % Small horizontal space between figures
    % Subfigure 2
    \begin{subfigure}[b]{0.2\textwidth}
        \includegraphics[width=1.1\textwidth]{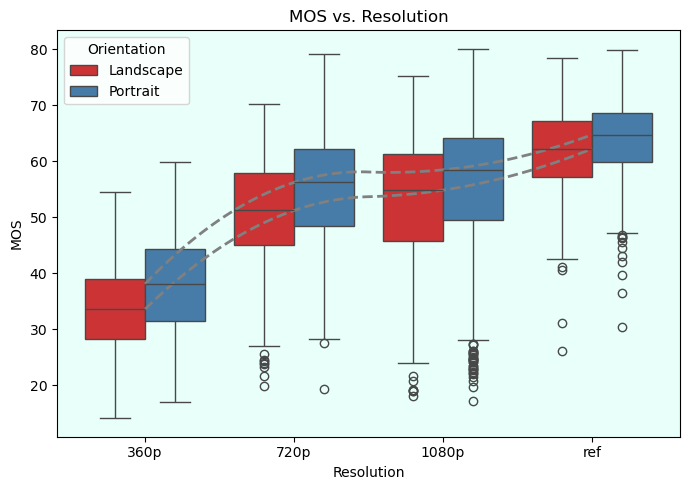}
        \caption{MOS vs. Resolution}
        \label{fig:mos_res}
    \end{subfigure}
    \hspace{0.5em} % Small horizontal space between figures
    \begin{comment}
        \begin{subfigure}[b]{0.2\linewidth}
        \centering
        \includegraphics[width=\textwidth]{images/mos-bitladder.png}
        \caption{MOS vs. Bitladder}
        \label{fig:mos_bitladder}
        \end{subfigure}
    \end{comment}
    \caption{MOS variations across bitrate, resolution, and combined bitladder.}
    \label{fig:bitladder_mos}
\end{figure}

\subsection{MOS Across Resolution and Bitrate}
Figure~\ref{fig:bitladder_mos} illustrates how resolution and bitrate influence perceptual quality. Fig.~\ref{fig:mos_bitrate} shows that MOS generally increases with bitrate, confirming that higher bitrates preserve visual quality by reducing compression artifacts. However, low-bitrate videos (0.2M, 0.5M) exhibit significant quality degradation, with MOS values spread across a wider range, indicating variability in perceptual impact depending on content complexity. Portrait videos tend to have slightly higher MOS across bitrate range. Fig.~\ref{fig:mos_res} highlights the direct correlation between resolution and perceived quality. As expected 360p videos receive the lowest MOS, while MOS steadily improves at 720p, and 1080p. The overlapping MOS distributions for 720p and 1080p suggest diminishing perceptual gains beyond a certain resolution threshold, influenced by content type and display scaling effects.

\subsection{Comparing CHUG with Existing HDR Datasets}
\begin{figure}[htb]
    \centering
    \includegraphics[width=0.7\linewidth]{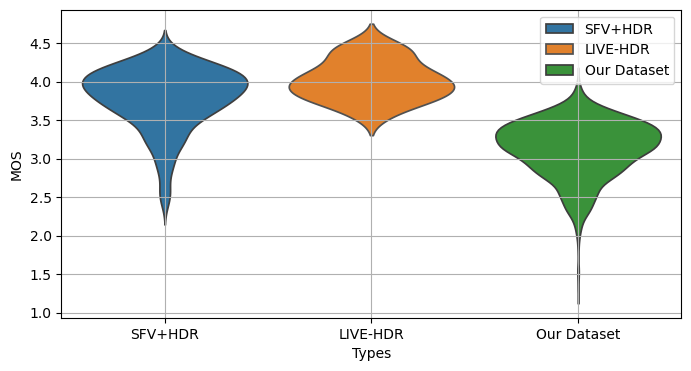}
    \caption{Comparison of MOS distributions: CHUG vs. LIVE-HDR vs. SFV+HDR.}
    \label{fig:dataset_comparison}
\end{figure}

Figure~\ref{fig:dataset_comparison} presents a comparison of MOS distributions across CHUG, LIVE-HDR, and SFV+HDR on common scale. LIVE-HDR\cite{livehdr} and SFV+HDR\cite{sfvhdr} exhibit a strong skew towards high MOS values, indicating a lack of diverse quality variations. These datasets primarily contain professionally generated or high-quality UGC, limiting their applicability in real-world HDR-VQA tasks. CHUG demonstrates a broader MOS distribution, covering low, medium, and high-quality HDR videos. This highlights its effectiveness in capturing diverse distortions and realistic UGC-HDR variations seen in modern streaming platforms. These findings reinforce the need for datasets like CHUG to bridge the gap between professional HDR content and real-world UGC-HDR challenges.

\begin{comment}
\begin{figure}[htbp]
    \centering
    \includegraphics[width=0.7\linewidth]{images/heatmap.png}
    \caption{Heatmap for bitladder over MOS. Same as MOS-Bitladder box plot.}
    \label{fig:overall}
\end{figure}
    
\end{comment}

%\vfill
%\pagebreak
\section{Conclusion}
\label{sec:conclusion}
In this paper, we introduced CHUG, a large-scale UGC-HDR video quality dataset with 5,992 videos and 211,848 subjective ratings collected via Amazon Mechanical Turk (AMT). Using SUREAL-based MOS computation, we ensured robust quality estimates. Our analysis highlights spatial-temporal complexity, orientation effects, and bitrate-resolution trade-offs in UGC-HDR videos. CHUG serves as a benchmark for No-Reference HDR-VQA models, reflecting real-world social media HDR content. The dataset and scores will be publicly available upon publication.

\bibliographystyle{IEEEbib}

\bibliography{main}

\end{document}